# Asynchronous Personalized Federated Learning through Global Memorization

Fan Wan[a], Yuchen Li[a], Xueqi Qiu[a], Rui Sun[b], Leyuan Zhang[a], Xingyu Miao[a], Tianyu Zhang[a], Haoran Duan[a] and Yang Long[a,*]

[a]*Department of Computer Science, Durham University, Durham, DH1,3LE, UK*
[b]*Department of Computer Science, Newcastle University, Newcastle, NE4 5TG, UK*

## ARTICLE INFO

## ABSTRACT

The proliferation of Internet of Things devices and advances in communication technology have unleashed an explosion of personal data, amplifying privacy concerns amid stringent regulations like GDPR and CCPA. Federated Learning (FL) offers a privacy-preserving solution by enabling collaborative model training across decentralized devices without centralizing sensitive data. However, statistical heterogeneity from non-independent and identically distributed datasets and system heterogeneity due to client dropouts—particularly those with monopolistic classes—severely degrade the global model's performance. To address these challenges, we propose the **Asynchronous Personalized Federated Learning (AP-FL)** framework, which empowers clients to develop personalized models using a server-side semantic generator. This generator, trained via **data-free knowledge transfer** under global model supervision, enhances client data diversity by producing both seen and unseen samples, the latter enabled by **Zero-Shot Learning** to mitigate dropout-induced data loss. To counter the risks of synthetic data impairing training, we introduce a **decoupled model interpolation method**, ensuring robust personalization. Extensive experiments demonstrate that AP-FL significantly outperforms state-of-the-art FL methods in tackling non-IID distributions and client dropouts, achieving superior accuracy and resilience across diverse real-world scenarios.

## 1. Introduction

The rapid proliferation of Internet of Things devices, from home automation systems and wearable health monitors to smart city sensors, coupled with the advances in communication technology, has led to an explosion of data generation in our daily lives. This vast expanse of data spans intricate applications such as facial recognition systems, detailed health data from fitness trackers, and extensive urban data from smart infrastructure, all of which harbor the potential to significantly advance the field of artificial intelligence. However, the deeply personal nature of such data, combined with an alarming escalation in privacy breaches, has intensified global scrutiny over data privacy. Legislative milestones like the European Union's General Data Protection Regulation (GDPR) and the California Consumer Privacy Act (CCPA) in the United States have underscored the imperative for robust data protection measures. These developments compel the AI community and scholars to innovate a framework that not only ensures rigorous protection of privacy but also enables efficient utilization of the burgeoning data, striking a crucial balance between utility and confidentiality.

In response to the urgent need for innovative solutions that preserve privacy while leveraging vast datasets, Federated Learning [1] has emerged as a groundbreaking paradigm. FL facilitates the collaborative training of a global model across multiple devices or clients without the necessity of centralizing local data. This decentralized approach involves each participant training models on their own devices, followed by the aggregation of these models into a cohesive global model, which is then updated and redistributed to all participants. By enabling data to remain securely on local devices, FL adeptly addresses the critical balance between data privacy and utility. Its application spans diverse sectors, from enhancing privacy in smart cities [2, 3] and improving diagnostic accuracy in healthcare [4, 5] to personalizing user experiences in digital services [6, 7], thereby illustrating its transformative potential in securely and efficiently harnessing data across industries.

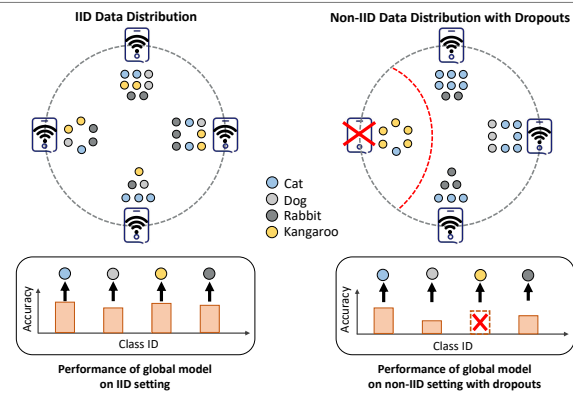

**Figure 1:** The illustration of the impact of non-IID data distribution and dropout clients with monopoly classes on global performance.

*Corresponding author.
E-mail: yang.long@durham.ac.uk (Y. Long)
ORCID(s): 0009-0005-1386-7847 (F. Wan)

Federated Learning, while promising, grapples with significant challenges such as statistical and systems heterogeneity. Statistical heterogeneity arises when data from diverse user devices vary widely due to factors such as geographical distribution, differing time zones, or unique user behaviors, leading to client drift. This phenomenon can degrade performance and slow the convergence of the global model, as seen in [8]. Furthermore, system heterogeneity compounds these issues, with disparities in device capabilities—like network bandwidth or battery life—affecting timely updates and further destabilizing training [9]. These heterogeneities not only challenge model training but also heighten the risk of creating monopolistic classes where single participants or groups disproportionately influence the model due to their unique data contributions.

The severe implications of monopolistic class dropouts, particularly within contexts of statistical and systems heterogeneity, are vividly illustrated in the healthcare sector. For example, if a healthcare provider uniquely treating a rare medical condition exits a federated network due to regulatory changes or technical failures, the global model instantly loses critical diagnostic data. This sudden dropout not only degrades the model's accuracy but also exposes the inherent vulnerabilities of relying on limited data sources. As depicted in Figure 1, while the model under idealized IID conditions might perform well, it encounters significant challenges in real-world settings marked by non-IID data distributions, especially when essential data sources vanish. This necessitates the development of innovative methods that effectively manage such dropouts, addressing both system heterogeneity and ensuring robust performance across diverse and realistic conditions.

Research on the impact of challenges from statistics and system heterogeneities have been extensive yet fragmented [10–12]. Previous studies [13–17] have addressed various dropout scenarios on the performance of global models, primarily under the assumption of independent and identically distributed (IID) conditions, where the impact of dropouts is minimal [13, 14, 18]. However, the real-world applicability of these findings is limited as they often overlook the complexities introduced by non-IID data distributions. Recent attempts to explore these issues in more realistic settings [15–17] have revealed significant gaps in existing methodologies, particularly in their ability to handle unpredicted dropouts and maintain data diversity without compromising the model's integrity.

In this research, we introduce the Asynchronous Personalized Federated Learning Framework (AP-FL) as a new approach to tackle the challenges of statistical and system heterogeneity. AP-FL employs a data-free knowledge transfer method to train a generator on the server side. With the aid of semantic information from Zero-Shot Learning and supervision from the received global model, the generator can generate seen samples from non-dropout clients and unseen samples from dropout clients to facilitate client model training. However, synthetic samples generated by the generator heavily rely on global model performance, which poses a risk when global model performance is suboptimal. To address this risk, we propose a decoupled model interpolation algorithm to mitigate the negative impact of synthetic data on Personalized model training. The main contributions of this work are summarized as follows:

- In order to address the non-IID challenge, we propose a novel personalized federated learning framework leveraging model interpolation.
- A novel FL framework to solve the class missing due to dropouts via data-free knowledge transfer and ZSL mechanism.

## 2. Related Work

**Statistic Heterogeneity** presents a major challenge in Federated Learning (FL) setups. Conventional FL approaches frequently experience client drift issues [8] in the presence of highly heterogeneous statistics (non-IID), resulting in diminished global model performance and suboptimal generalization across numerous clients. To address this challenge, several existing works [10–12, 19–22] have started to research Personalized Federated Learning (PFL) which has recently gained considerable attention for its ability to adapt the global model to better fit each client's local data distribution. One of the research methodologies focuses on personalized a single global model by introducing techniques, such as Data Augmentation[23], Client Selection[24], Regularization[9], and Meta-Learning[25].

Data augmentation, such as FAug [23], promotes statistical homogeneity by generating new data or using proxy data for clients, enabling the satisfaction of the IID assumption and benefiting the training of a unified global model through server-side generative adversarial networks trained with limited client-side samples for IID dataset generation. Client selection, such as the adaptive reinforcement learning algorithm proposed by [24], identifies representative client subsets to capture the global data distribution, mitigating non-IID data impact and improving the performance and communication efficiency of the trained model. Model regularization, exemplified by Fedprox[9], introduces a regularization term in the loss function to constrain personalized models from deviating significantly from the global model, effectively limiting the impact of irregular client updates. Meta-learning, inspired by local fine-tuning from the global model, was introduced into PFL, building initial meta-models for clients to fine-tune after one model gradient descent step, as exemplified by [25], which combines meta-learning and reinforcement learning to adaptively optimize the federated learning process.

**Systems Heterogeneity** as another crucial factor beyond statistical heterogeneity that should be considered in the federated network, since interplay exists between them in federated learning [26]. In a real-world federated training task, thousands of devices possibly participate, with diverse system-level attributes, hardware configuration(CPU, Memory), network connectivities (wired and wireless network), and battery capability [26, 27]. Such characteristics substantially heighten the uncertainty within a federated network, giving rise to challenges such as misleading optimization direction, straggler issues, and client dropout problems. To tackle systems and statistical heterogeneity problems, [27] proposed an adaptive client sampling algorithm that reduces convergence duration by determining the relationship between overall learning time and sampling probabilities. In addition, [26] proposed a novel federated optimization algorithm, widely known as FedProx. FedProx alleviates the impact of systems and statistical heterogeneity on convergence behavior by introducing a proximal term to the objective, thereby increasing stability. This addition offers a principled approach to handling heterogeneity associated with partial information, allowing for convergence guarantees and an analysis of the effects of heterogeneity. While these approaches have effectively mitigated the impact of system and statistical heterogeneity issues broadly, their efficiency remains limited in specific situations, such as client dropout problems. Most recently, very limited studies have started focusing on client drop problems. [28] propose the concept of "friendship" between clients, wherein clients with similar data distributions and local model updates are considered friends. This approach seeks to alleviate the impact of client dropout by substituting a friend client's local model update for the dropout client's update when computing the next round global model, resulting in minimal substitution error. However, while this method mitigates the negative impact on global model performance, it does not enhance the global model's effectiveness on the dropped client's local dataset.

**Asynchronous Personalized Federated Learning.** Building on the advancements in Personalized Federated Learning (PFL), our proposed Asynchronous Personalized Federated Learning (AP-FL) framework is designed to address both statistical and system heterogeneity. In PFL, the challenge of non-IID data is tackled by allowing each client to develop a model that is personalized to its local data distribution. In AP-FL, this personalization is achieved by maintaining the same model architecture across clients but allowing for distinct model weights that adapt to the specific data characteristics of each client. This approach ensures that while all clients benefit from shared global knowledge, their individual models are fine-tuned to address local data heterogeneity. To mitigate the effects of client dropouts, AP-FL incorporates a novel data-free knowledge transfer mechanism, allowing the generation of synthetic samples that aid in the continuous training of client models even when clients are temporarily offline. This strategy effectively handles both asynchronous updates and the challenges posed by varying client availability, leading to a more robust and efficient federated learning process.

**Zero-Shot Learning.** Zero-Shot Learning (ZSL) [29–32] has become a widely used approach in deep learning for recognizing unknown or unseen classes by leveraging the relationship between seen and unseen classes via class semantic information. In recent years, various works have been proposed to address this challenge, such as building the mapping between visual and semantic space or generating unseen class data to alleviate the issue of missing data. In the test phase, conventional ZSL (CZSL) [33] methods assume that test data only come from unseen classes, while generalized ZSL (GZSL) [34] assigns both seen and unseen data to corresponding classes. Building on the pioneering work of [30], we propose a data-free method to train a semantic generator capable of generating synthetic samples from seen and unseen data separately. This approach enables all clients to develop personalized models asynchronously with the help of a semantic generator, regardless of the distribution of their data.

## 3. Methodology

This section presents the proposed AP-FL. We first describe the problem statement, followed by AP-FL framework design. Several key modules of AP-FL are detailed at both the server and client sides.

### 3.1. Problem Statement

Conventional federated learning approaches, such as FedAvg [1], address $C$-class classification problems across $K$ clients. For each client $k \in \{1, 2, \ldots, K\}$, its private local dataset $\mathcal{D}_k$ is drawn from the local data distribution $p_k(x, y)$, where $x \in \mathcal{X}$ is the input feature, and $y \in \mathcal{Y}$ denotes the corresponding label. The goal of FL is to enable clients to jointly train a global model with parameters $\theta^*$ over the combined global dataset $\mathcal{D} = \bigcup_k \mathcal{D}_k$.

The global objective is to find the optimal model parameters $\theta^*$ that minimize the global loss $\mathcal{L}(\theta)$, which can be formulated as:

$$\theta^* = \arg\min_\theta \mathcal{L}(\theta), \tag{1}$$

where $\mathcal{L}(\theta)$ represents the empirical loss over the entire global dataset $\mathcal{D}$. This global loss is computed as the weighted sum of the local losses $\mathcal{L}_k(\theta)$ from each client $k$, with the weights proportional to the size of each local dataset $\mathcal{D}_k$:

$$\mathcal{L}(\theta) = \sum_{k=1}^{K} \frac{|\mathcal{D}_k|}{|\mathcal{D}|} \mathcal{L}_k(\theta), \tag{2}$$

here, $\mathcal{L}_k(\theta)$ denotes the local loss for client $k$, and $\frac{|\mathcal{D}_k|}{|\mathcal{D}|}$ is the weighting factor based on the proportion of data that client $k$ contributes to the global dataset.

All clients aim to optimize the global model $\theta$ by minimizing their local expected risk:

$$\mathcal{L}_k(\theta) = \mathrm{E}_{(x,y) \in \mathcal{D}_k} \mathcal{L}(\theta; (x, y)). \tag{3}$$

The key steps involved in a complete FL training process are outlined below: (i) At communication round $t$, the aggregator server randomly selects $K$ clients available for training and sends the global model $\theta^*$ to the selected clients, which they deploy as a local model, $\theta_k^t$. (ii) Each selected client trains its local model $\theta_k^t$ using its dataset $\mathcal{D}_k$ for $E$ local epochs. (iii) Once the aggregator server collects local model updates from enough participants, $\theta_k^{t+1}$, the server aggregates all updates based on Equation 2. (iv) Repeat steps (i)~(iii) until the model reaches convergence.

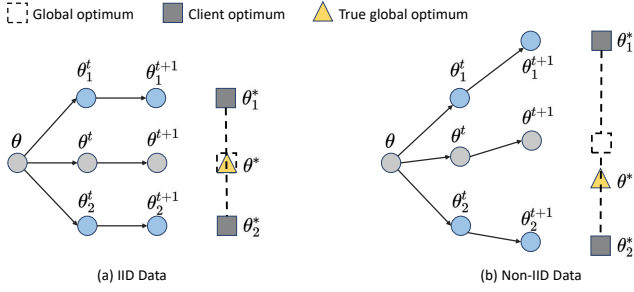

**Figure 2:** Illustration of client drift in FedAvg in Dirichlet non-IID settings.

Client drift issues posed a serious challenge when implementing FL in the real world. The performance and efficacy of the vanilla FedAvg algorithm have been demonstrated in Independent and Identically Distributed (IID) settings, where each client has similar data distribution, and samples are identically distributed among clients. However, it fails in Non-IID settings, where data distribution between clients can be highly skewed, and sample distribution may differ significantly. This can lead the locally trained model to be optimized in a direction that deviates significantly from its trained in an IID dataset.

Figure 2 illustrates how FedAvg performs in both IID and non-IID settings. The average model $\theta^{t+1}$ is equidistant to both local optima $\theta_1^*$ and $\theta_2^*$ in an IID setting, which brings it closer to the global optimum $\theta^*$. However, in Non-IID settings, the resulting average model $\theta^{t+1}$ may not be close to the global optimum $\theta^*$, causing the global model not to converge to its true global optimum. In these scenarios, the single global model is difficult to generalize well to all clients, and the performance of the global model may not even exceed the local model where the client does not participate in FL training[35]. This is contrary to the original intention of the client to participate in FL.

Analogous to the straggler issue in distributed systems, client dropout is a prevalent phenomenon in federated networks with system heterogeneity. In certain non-IID scenarios, such as those characterized by extreme shifts in data quantity and class categories, client dropout can amplify the adverse effects on global model optimization. An existing study [35] indicates that, given a sufficient number of clients continuously participating in federated learning training under IID data settings, the accuracy of the global model remains unimpaired, even if permanent dropouts among some clients. However, in non-IID scenarios, where some clients have unique or minority classes that are not present in the datasets of other clients, the dropout of those clients can significantly negatively impact the performance of the global model. This is because the performance of the global model relies on contributions from all participating clients to learn a representative model. When a dropout client has unique class category that is not represented by other clients, as the training continues, the global model will be fitted to the optimal of the other available classes, resulting in an extremely rapid decline in the global model's ability to identify the missing class data.

Our work is motivated by recent advances in PFL[36], but it goes beyond it by addressing system heterogeneity, specifically the challenge of client dropout, in addition to the problem of statistical heterogeneity. We aim to develop a global knowledge that can help non-dropout and dropout clients to build a personalized model that can tackle client local drift issues, even when the data on dropout clients are distinct from those on all non-dropout clients. To achieve this goal, we propose to train a personalized supervised classification model for a group of non-dropout clients $S_n$ and dropout clients $S_d$, where $S_n, S_d \in K$.

$$\theta_k^p = \arg\min_{\theta_1,\ldots,\theta_K} \sum_{k \in S_n \cup S_d} \frac{|\mathcal{D}_k|}{|\mathcal{D}|} \mathcal{L}_k(\theta_k^p), \quad (4)$$

where $\theta_k^p$ represents the personalized model residing on client $k$. Our approach is distinct from other methods that aim to mitigate client dropout, as our focus is not only on dropout clients but on all clients. By enabling dropout clients to benefit from global knowledge and establish their personalized models, our method can address the challenge of statistical heterogeneity while also tackling the issue of client dropout.

### 3.2. Proposed Framework: AP-FL

Numerous studies in recent years have focused on addressing statistic and systems heterogeneity by capturing global knowledge, such as GAN-based approaches [37–41]. However, most of them require the generator access to clients' raw data, contradicting the original principles of federated learning. Alternatively, knowledge distillation-based methods [42–44] rely on a proxy dataset and tackle client drift issues by leveraging disagreement between global and client models. Nevertheless, the availability of a proxy dataset in real-world federated learning scenarios cannot always be guaranteed.

To tackle these challenges posed by client drift and client dropout in above non-IID scenarios, we introduce a novel federated learning framework termed AP-FL, illustrated in Figure 3. AP-FL is a plugin that could cap into most widely use neural network, and features a lightweight semantic generator, maintained by the central server, which captures global knowledge through data-free knowledge transfer from the global model. This semantic generator is disseminated to non-dropout clients to support the development of personalized models tailored to their data distribution. Considering the likelihood of a single client dominating minority classes in real applications, we adopt the Zero-Shot learning paradigm, enabling the semantic generator to create synthetic data for minority classes present on dropout clients.

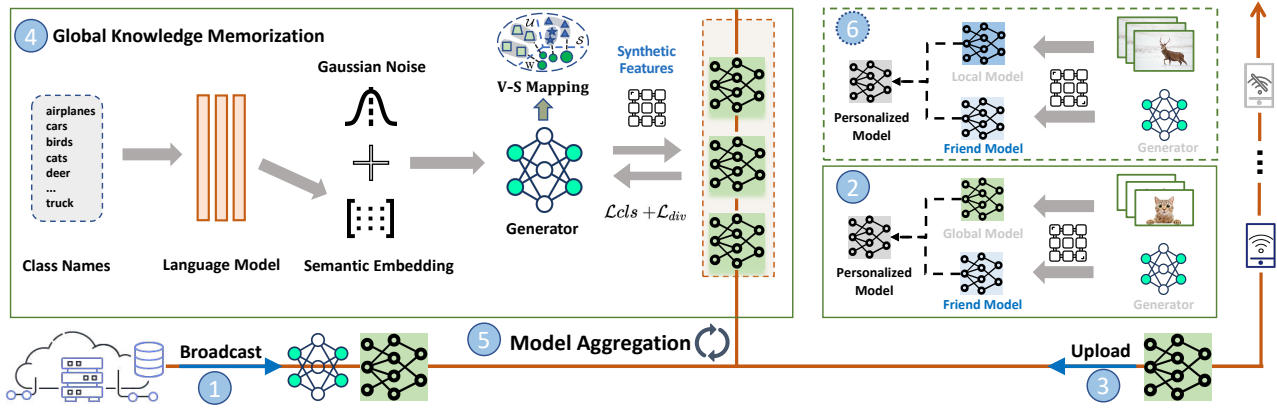

**Figure 3**: Overview of the Asynchronous Personalized Federated Learning.

This is achieved by establishing a mapping between semantic information and features, even without direct access to the dropout client data by the global model. Consequently, this approach facilitates asynchronous training of personalized models by dropout clients based on their unique data distribution, supported by the semantic generator.

**Global Knowledge Memorization.** As previously discussed, non-IID scenarios can result in client drift issues, adversely affecting model performance. Therefore, it is essential to devise a conditional generator, denoted as $G$, maintained on the central server side to capture the global perspective of data distribution. This generator aims to assist each client in developing a personalized model $\theta_k$ while preserving user privacy. Specifically, the server broadcasts $G$ to support non-dropout (non-dropout) clients $S_n$ in training personalized models by generating synthetic samples that enhance the diversity of client data distribution. The completed process of global knowledge memorization could be summarized as follows: Firstly, the generator is initialized on the server-side as follows:

$$\hat{x} = G(z, y; \omega), \quad (5)$$

here, $\omega$ denotes the parameters of $G$, and $z \sim \mathcal{N}(\mathbf{0}, \mathbf{1})$ represents the standard Gaussian noise, which is introduced to increase the diversity of the generated data and reduce overfitting. The variable $y$ is the label representing the desired output class, while $\hat{x}$ is the synthetic sample corresponding to the input noise $z$ and label $y$.

Due to the scarcity of resources for training $G$, only the global model $\theta^*$ and the client local models $\theta_k$ are accessible. Therefore, it is imperative to ensure that the synthetic samples $\hat{x}$ generated by the $G$ are compatible with the input space of client local models $\theta_k$. This can be formulated as follows:

$$\mathcal{L}_{ce} = -\sum_{i=1}^{C} y_i \log\left(\sigma\left(D\left(\hat{x}; \theta_k\right)_i\right)\right), \quad (6)$$

here, $i$ indexes the classes, and $C$ represents the total number of classes. The softmax function $\sigma(\cdot)$ outputs a probability distribution over the $C$ classes, and $D(\hat{x}; \theta_k)$ denotes the output of the client model $\theta_k$ when given the synthetic sample $\hat{x}$. The term $y_i$ is the ground truth label for class $i$, and the cross-entropy loss $\mathcal{L}_{ce}$ measures how well the model's predicted distribution aligns with the true labels.

To well fit the synthetic samples effectively with each client model's data distribution, we incorporate a weighted average of the loss function, considering the distribution of distinct categories for each user. Consequently, the weighted average cross-entropy loss is defined as follows:

$$\mathcal{L}_{cls} = \sum_{k \in S_o} \alpha_k^y \mathcal{L}_{ce}^k, \quad (7)$$

where $\alpha_k^y$ represents the proportion of samples in class $y$ of the $k$-th non-dropout client in the entire global training set, and $\mathcal{L}_{ce}^k$ represents the cross-entropy loss produced by $k$-th non-dropout client.

Employing only the $\mathcal{L}_{cls}$ may result in the generator's model collapse [45], causing $G$ to output identical data for every class. To motivate $G$ to enhance the diversity of synthetic samples, we incorporate a regularization term into the loss function. Specifically, we introduce a diversity loss term, which encourages the generator to generate varied samples. The diversity loss is defined as follows:

$$\mathcal{L}_{diversity} = -\frac{1}{n_s} \sum_{i=1}^{n_s} \sum_{j=1, j \neq i}^{n_s} \frac{\left|\hat{x}_i - \hat{x}_j\right|_2}{n_s - 1}, \quad (8)$$

where $n_s$ is the number of synthetic samples, and $\hat{x}_i$ and $\hat{x}_j$ are two different synthetic samples of same classes generated by the generator $G$. The term $n_s - 1$ in the denominator is a normalization factor to appropriately scale the diversity loss. This loss term encourages the generator to produce diverse synthetic samples by minimizing the Euclidean distance between any two different synthetic samples. The overall loss function for the generator is then defined as follows:

$$\mathcal{L}_G = \lambda \mathcal{L}_{cls} + (1 - \lambda) \mathcal{L}_{diversity}, \quad (9)$$

where $\lambda$ is the hyper-parameters that control the relative importance of the two loss terms. By minimizing this loss function, the generator $G$ is encouraged to produce diverse synthetic samples that better capture the underlying data distribution of the client models.

**PFL via Decoupled Model Interpolation.** The majority of existing studies addressing non-IID problems concentrate on data generation-based methods, such as mixing up non-IID real and synthetic data into a unified IID training set for each client's local model or utilizing fake data to capture the disagreement between global and local models for bi-level knowledge distillation, ultimately enhancing the performance of global or local models. However, for both approaches, the data generation capability of the generator heavily relies on the global model's accuracy. Consequently, the training quality of the generator cannot be guaranteed in this manner, a limitation similar to that encountered in our solution.

To overcome this challenge, we propose a decoupled model interpolation method that modulates the impact of synthetic samples within personalized federated learning. In this approach, users utilize the trained generator to generate synthetic samples $\hat{x}$ conforming to their local data distribution $\mathcal{P}_k$. These synthetic samples are subsequently employed to train a classifier, referred to as the friend model. Finally, we combine the client model and friend model to create a personalized model that more effectively adapts to the user's local data. The following equation illustrates the decoupled model interpolation method:

$$\theta_k^p = \beta\theta_k + (1-\beta)\theta_k^f, \tag{10}$$

where $\theta_k^f$ and $\theta_k^p$ represent the model parameters of the friend model and personalized model separately in $k$-client, and $\beta$ represents the confidence coefficient for the friend model.

**PFL for the dropouts.** In addition to tackling statistic heterogeneity, AP-FL also handles client dropout issues lead by systems heterogeneous. In real-world FL training, which may involve thousands of clients, communication bandwidth constraints within a distributed system necessitate the selection of only a limited number of clients to participate in each training round. This situation can result in clients possessing all data of minority classes not engaging in FL training from start to finish before dropping out. This implies that those data categories are never present in any non-dropout client, and we refer to them as unseen classes for the global model. Sending the global model to these dropouts would be unproductive, as the global model has not seen data from these dropouts' categories and, consequently, cannot identify the data for these categories.

Inspired by the works in Zero-Shot Learning [46], we distinguish data from non-dropouts and dropouts as seen data $\mathcal{D}_s$ and unseen data $\mathcal{D}_u$, respectively. The relationship between their data categories is disjoint and can be formulated as $\mathcal{Y}_s \cap \mathcal{Y}_u = \emptyset$. The main challenge lies in obtaining informative semantic information that allows the generator to establish the mapping between features and semantic information, enabling the generator to synthesize unseen data from dropout clients. Conventional ZSL approaches benefit from auxiliary semantic embedding information, such as attributes annotated by experts in relevant fields. However, traditional FL datasets lack such auxiliary information. To tackle this issue, we employ foundation models like BERT [47] and CLIP [48], which are pre-trained on extensive data and can predict underlying properties, such as attributes.

While large foundation models such as BERT and CLIP have been leveraged to aid the semantic embedding process, it is important to acknowledge that these models introduce strong priors due to their extensive pre-training on large-scale datasets. This could potentially raise concerns about the fairness of comparison with other approaches that do not utilize such pre-trained models. To mitigate this, we carefully ensure that the usage of BERT and CLIP is balanced with techniques that limit their overwhelming influence on final model performance. Additionally, their role is primarily to provide semantic structure in the absence of labeled data, rather than directly contributing to model learning in traditional ways. By using BERT and CLIP in a controlled manner, we aim to ensure that the comparison with other FL systems remains as fair and unbiased as possible, focusing on the federated learning performance rather than overreliance on pre-trained representations.

To develop the mapping between features and semantic information, we support the Generator on the central server side, which can generate pseudo data from $\mathcal{D}_s$ to $\mathcal{D}_u$. So the input of Generator in Eq. (5) should become the following format:

$$\hat{x} = G(z, A(y); \theta), \tag{11}$$

where the $A(\cdot)$ represents auxiliary semantic embedding. Finally, the personalized model in all clients can be formulated as follows:

$$\theta_k^p = \begin{cases} \beta\theta_k + (1-\beta)\theta_k^f & \text{non-dropout clients;} \\ \beta\theta_k^l + (1-\beta)\theta_k^f & \text{dropout clients.} \end{cases} \tag{12}$$

Where $\theta_k^l$ represents the localized global model for the dropout in $k$-client, and $\theta_k^f$ and $\theta_k^p$ denote the model parameters of the friend model and personalized model separately in $k$-client.

The global knowledge model captures rare class distributions through the ZSL approach. When dropout clients contain rare or unique class data, the generator synthesizes samples for these classes using semantic embeddings. This approach helps maintain model performance in non-IID settings by generating synthetic data for unseen or underrepresented classes. By mapping semantic features to the output space, the model can generalize across diverse data distributions, ensuring that rare classes are effectively learned. As a result, both dropout and non-dropout clients benefit from accurate and robust personalized models, ensuring comprehensive generalization across all classes.

**Discussion.** Our proposed Personalized Federated Learning (PFL) approach represents an innovative adaptation of Clustering-based Federated Learning (CFL) principles, specifically designed for client-side implementation. In CFL, clients with similar data distributions are grouped together to collaboratively train a shared model, enabling mutual learning among "friend" models within the same cluster. However, the dynamic grouping mechanism in CFL can become unstable when new data samples emerge, as these may shift client distributions and disrupt the training process. To address this, our approach introduces a more flexible and adaptive strategy by leveraging synthetic data generation. Clients can continuously generate synthetic samples that reflect their evolving data distributions, allowing them to train personalized models using local clustering techniques. These personalized models are finely tuned to each client's unique data characteristics, resulting in significantly enhanced generalization performance.

Additionally, our framework incorporates asynchronous aggregation to bolster the robustness of the training process, especially in challenging environments marked by client dropouts or system heterogeneity. With asynchronous aggregation, the global model is updated immediately upon receiving updates from any client, minimizing latency and boosting training efficiency. However, we acknowledge that this approach may introduce consistency challenges, as updates could be derived from stale or outdated models. This trade-off differs from synchronous aggregation, which ensures consistent global model updates by waiting for contributions from all clients but risks delays due to slower or offline participants. In our method, asynchronous aggregation effectively mitigates the effects of client dropouts and system variability, while maintaining high model performance through carefully crafted update mechanisms.

## 4. Experiments

In this section, we present the evaluation of the effectiveness of our proposed method, AP-FL, and compare it with several advanced methods in different datasets and settings. The evaluation focuses on two key aspects: (1) personalized model accuracy in non-dropout clients, and (2) the improvement in model accuracy for dropout clients with the assistance of global knowledge.

### 4.1. Basic Set

**Dataset:** This study presents experimental results on four diverse image datasets: CIFAR10, CIFAR100 [49], EMNIST [50], and Fashion MNIST [51]. The CIFAR10 dataset contains 60,000 32x32 color images divided into ten classes, which has been widely used for image classification tasks. CIFAR100 is a more challenging variant of CIFAR10, consisting of 100 classes. The EMNIST dataset is a collection of over 800,000 images of 26 handwritten letters, while Fashion MNIST comprises 70,000 grayscale images of 28x28 pixels, representing ten different clothing categories. To maintain consistency in image resolution, we resized all images to 32x32 pixels. To evaluate our model, we set aside 10% of the data for testing purposes, and we distributed the test data among the clients while ensuring that the test data had the same label distribution as the training data on each client's side.

**Table 1**
Data Partitioning for $\gamma = 2$ Pathological Non-IID on CIFAR10 dataset, in the Dropout Setting. The classes [8, 9] denote the minority classes monopolized by rare clients.

| Device No. | 0 | 1 | 2 | 3 | 4 |
|---|---|---|---|---|---|
| Classes | 0, 1 | 2, 3 | 6, 7 | 4, 5 | 2, 4 |
| Device No. | 5 | 6 | 7 | 8 | 9 |
| Classes | 2, 3 | 6, 7 | 4, 5 | [8, 9] | 0, 1 |

**Heterogeneity Settings:** The performance of the proposed AP-FL framework is evaluated in two distinct heterogeneity settings to analyze its efficacy under varying degrees of heterogeneity. (1) **Full Participated Setting**, solely accounts for statistical heterogeneity and considers an ideal FL scenario where all clients are available and selected randomly by the server without dropped calls. Similar to [52, 53], we adopt the Dirichlet Distribution $Dir(\alpha)$ to control the degree of non-IID distribution. Specially, we set $\alpha$ to three different values, namely 0.1, 0.05, and 0.01, across three image datasets - CIFAR10, CIFAR100, and EMNIST. Since FEMNIST already considers various kinds of imbalances, such as data heterogeneity, data imbalance, and class imbalance, we did not apply the Dirichlet distribution to FEMNIST. Furthermore, we varied the number of clients to five and ten to simulate different levels of non-IID data. (2) **Dropout Setting**, a dropout factor is introduced to simulate more practical scenarios where FL training encounters both statistical and system heterogeneity. In this setting, we adopt the Pathological non-IID[54] approach, where only certain classes of data are assigned to each client. We simulate ten clients to jointly train a global model in all datasets, and then we use the hyper-parameter $\gamma$ to control the number of classes on each client. As shown in Table 1, when $\gamma = 2$ means that there are two classes of data on each client. We assume some rare clients with monopoly classes will drop out to verify the effectiveness of the proposed personalized model in dropout clients.

**Baselines:** This study presents a comprehensive comparison of the proposed AP-FL framework with several baseline algorithms in two distinct settings. **In the Full Participated Setting**, we compare AP-FL against FedAvg [55], FedProx [9], SCAFFOLD [8], FedGen [45], and FedDF [56]. In addition, we evaluate the performance of AP-FL against local training, which involves training a local model without the use of federated learning. **In the Dropout Setting**, we compare FedAvg [55] and local training as the baseline approaches. For FedAvg, we conduct one-off fine-tuning training for the global model trained by the non-dropout client in the dropout client with its monopoly classes and then test its global model performance in monopoly classes. For local training, we send the initial global model to dropout clients and train the local model without federated learning.

**Implement Details:** We implement all experiments of AP-FL in PyTorch, where the classifier in all experiments is a standard CNN model, which consisting of two 5 × 5 convolution layers (the first with 32 channels, the second with 64 channels, each followed with 2 × 2 max polling), two fully connected layers each with 1600, 512 units and ReLU activation. For semantic embedding, we use 512-dimensional word embedding generated by CLIP [48]. Our generator network architecture is borrowed from [57], but we replace the input of an original one-hot label with the semantic embedding generated from various models. All methods were trained with a batch size of 50 and optimized using the Adam optimizer with an initial learning rate of 0.0002, for a total of 20 local training epochs. During the generator training stage, synthetic samples of size 600 for each class were fed into each non-dropout client model to supervise the generator training. The hyperparameter $\lambda$ was set to 0.5 for each dataset, and the server aggregated the loss from different client models based on the proportion of samples in the classes of each client. Finally, the trained generator and aggregated global model were broadcasted to each client to complete personalized model training, using a hyperparameter of $\beta = 0.01$ for CIFAR10 and CIFAR100 and $\beta = 0.1$ for EMNIST and Fashion MNIST datasets.

**Table 2**
Comparison with SOTA FL algorithms in Full Participation settings

| Dataset | Client Num | Heteroge. Setting | Test Accuracy(%) | | | | | | |
|---|---|---|---|---|---|---|---|---|---|
| | | | Local | FedAvg | FedProx | SCAFFOLD | FedGen | FedDF | AP-FL |
| CIFAR10 | 5 | $\alpha = 0.01$ | 15.71 ± 0.39 | 43.83 ± 0.90 | 51.48 ± 1.21 | 54.47 ± 0.99 | 28.66 ± 1.19 | 44.66 ± 1.40 | **61.84 ± 1.75** |
| | | $\alpha = 0.05$ | 28.72 ± 0.32 | 61.61 ± 1.36 | 60.08 ± 3.19 | 64.28 ± 1.43 | 41.86 ± 0.47 | 60.27 ± 0.39 | **65.14 ± 0.32** |
| | | $\alpha = 0.1$ | 33.00 ± 1.16 | 65.77 ± 1.77 | 65.07 ± 0.40 | 67.37 ± 1.02 | 46.61 ± 2.88 | 64.58 ± 0.95 | **69.46 ± 0.18** |
| | 10 | $\alpha = 0.01$ | 15.76 ± 0.04 | 38.79 ± 4.97 | 45.98 ± 0.58 | 46.09 ± 2.50 | 26.67 ± 2.50 | 37.06 ± 1.26 | **56.28 ± 0.51** |
| | | $\alpha = 0.05$ | 24.95 ± 0.87 | 52.96 ± 0.24 | 51.68 ± 0.32 | 53.01 ± 0.74 | 27.51 ± 1.76 | 52.07 ± 1.97 | **58.73 ± 1.75** |
| | | $\alpha = 0.1$ | 35.04 ± 1.54 | 58.15 ± 0.94 | 56.36 ± 0.26 | 60.04 ± 1.08 | 43.08 ± 0.55 | 57.89 ± 1.00 | **61.39 ± 0.28** |
| CIFAR100 | 5 | $\alpha = 0.01$ | 13.89 ± 0.34 | 30.16 ± 0.42 | 29.28 ± 0.13 | 33.80 ± 1.19 | 30.04 ± 2.14 | 30.47 ± 1.43 | **35.28 ± 4.21** |
| | | $\alpha = 0.05$ | 24.53 ± 0.44 | 32.19 ± 2.13 | 34.58 ± 1.05 | 36.74 ± 0.41 | 32.17 ± 1.21 | 35.34 ± 1.32 | **38.47 ± 0.42** |
| | | $\alpha = 0.1$ | 25.23 ± 0.38 | 34.63 ± 0.32 | 34.89 ± 0.49 | 37.18 ± 1.73 | 34.93 ± 1.03 | 36.84 ± 2.41 | **39.95 ± 1.45** |
| | 10 | $\alpha = 0.01$ | 14.47 ± 1.53 | 28.37 ± 1.10 | 28.11 ± 1.03 | 30.32 ± 1.05 | 28.18 ± 0.58 | 28.39 ± 2.65 | **31.74 ± 1.52** |
| | | $\alpha = 0.05$ | 23.40 ± 0.28 | 30.01 ± 0.56 | 32.16 ± 0.50 | 33.49 ± 0.73 | 29.55 ± 0.41 | 33.12 ± 1.74 | **35.86 ± 0.47** |
| | | $\alpha = 0.1$ | 24.09 ± 1.53 | 32.34 ± 0.65 | 32.78 ± 0.13 | 34.95 ± 0.58 | 31.88 ± 0.65 | 33.51 ± 1.24 | **36.74 ± 0.44** |
| EMNIST | 5 | $\alpha = 0.01$ | 24.36 ± 0.23 | 86.56 ± 0.95 | 85.43 ± 0.61 | 85.30 ± 0.37 | 82.41 ± 2.34 | 88.06 ± 0.37 | **89.07 ± 1.26** |
| | | $\alpha = 0.05$ | 33.20 ± 0.29 | 89.33 ± 0.16 | 87.97 ± 0.40 | 89.22 ± 0.21 | 86.86 ± 0.89 | 89.27 ± 0.27 | **91.24 ± 0.52** |
| | | $\alpha = 0.1$ | 36.86 ± 0.26 | 90.85 ± 0.31 | 89.36 ± 0.55 | **91.88 ± 0.46** | 90.12 ± 0.63 | 90.32 ± 0.26 | 91.60 ± 0.16 |
| | 10 | $\alpha = 0.01$ | 13.38 ± 0.26 | 65.98 ± 3.95 | 77.09 ± 1.49 | 69.23 ± 1.47 | 66.74 ± 8.45 | 65.72 ± 1.33 | **82.48 ± 0.43** |
| | | $\alpha = 0.05$ | 19.03 ± 0.03 | 82.32 ± 0.35 | 83.23 ± 0.71 | 84.06 ± 1.24 | 81.05 ± 1.69 | 83.19 ± 1.27 | **85.27 ± 0.16** |
| | | $\alpha = 0.1$ | 32.22 ± 0.02 | 88.69 ± 0.47 | 87.68 ± 0.47 | 87.88 ± 0.81 | 88.45 ± 0.49 | **89.12 ± 0.16** | 88.94 ± 1.20 |
| Fashion MNIST | 5 | - | 49.15 ± 0.19 | 88.28 ± 0.89 | 87.68 ± 0.89 | 88.60 ± 1.20 | 87.05 ± 2.21 | 88.79 ± 0.95 | **89.36 ± 0.58** |
| | 10 | - | 41.61 ± 0.73 | 85.94 ± 1.51 | 85.74 ± 0.16 | 85.50 ± 0.45 | 85.23 ± 2.44 | 83.97 ± 3.62 | **87.04 ± 0.17** |

## 4.2. Experimental Results

**Comparison with SOTA in Full Participated Settings:** Table 2 presents a comprehensive evaluation of the accuracy of various algorithms on different Dirichlet non-IID distributions, demonstrating that our proposed AP-FL framework surpasses most state-of-the-art (SOTA) methods, particularly in highly heterogeneous scenarios, such as alpha = 0.01 or 0.05. Furthermore, we increased the skewness of label distribution between clients by expanding the number of clients. As Table 1 demonstrates, even in this 10-client scenario, AP-FL maintains superior performance over other algorithms. Compared to FedGen, which directly feeds synthetic data into the global model, our approach can effectively alleviate the impact of spurious data on model performance through the decoupled model interpolation technique. Additionally, Figure 4 shows the comparative performance of all algorithms at varying degrees of label distribution skewness, with AP-FL demonstrating more consistent and stable performance as data heterogeneity increases.

**Comparison with Existing Works in Dropout Settings.** Table 3 presents the results of the comparison between our proposed AP-FL framework and the Local and FedAvg-FT baselines. Our Personalized Model trained with AP-FL on CIFAR10, EMNIST, and Fashion MNIST datasets outperforms the Local model and FedAvg-FT in most cases, indicating the effectiveness of our approach in assisting dropout clients to train their own Personalized model. However, the performance of AP-FL on CIFAR100 is slightly behind FedAvg-FT. We attribute this to the fine-grained nature of the dataset, which poses a challenge for language models like CLIP/BERT to generate semantically distinctive information for subclasses under certain categories, leading to poor quality of generated unseen synthetic samples. In summary, our findings suggest that when clients with monopolistic categories drop out, AP-FL presents a more competitive alternative to training a local model or fine-tuning a global model for those dropout clients.

**Table 3**
Comparison with FedAvg in Dropout settings. 'MC' represent the missing classes due to dropout client with minority classes.

| Dataset | CIFAR10 | | CIFAR100 | | EMNIST | | Fashion MNIST | |
|---|---|---|---|---|---|---|---|---|
| MC(%) | 10% | 20% | 10% | 20% | 10% | 20% | 10% | 20% |
| Local | 29.47 ± 1.69 | 26.74 ± 0.49 | 22.61 ± 1.52 | 21.97 ± 1.10 | 30.15 ± 2.14 | 29.73 ± 1.19 | 47.51 ± 1.06 | 47.63 ± 0.98 |
| FedAvg-FT | 31.43 ± 0.58 | 29.82 ± 1.19 | **23.15 ± 1.32** | **24.73 ± 1.43** | 34.81 ± 2.41 | 34.05 ± 0.56 | 51.78 ± 0.37 | 51.96 ± 0.74 |
| AP-FL | **34.18 ± 0.49** | **32.97 ± 0.26** | 23.12 ± 1.55 | 24.65 ± 1.08 | **37.91 ± 0.71** | **36.29 ± 0.28** | **58.97 ± 0.58** | **56.83 ± 0.32** |

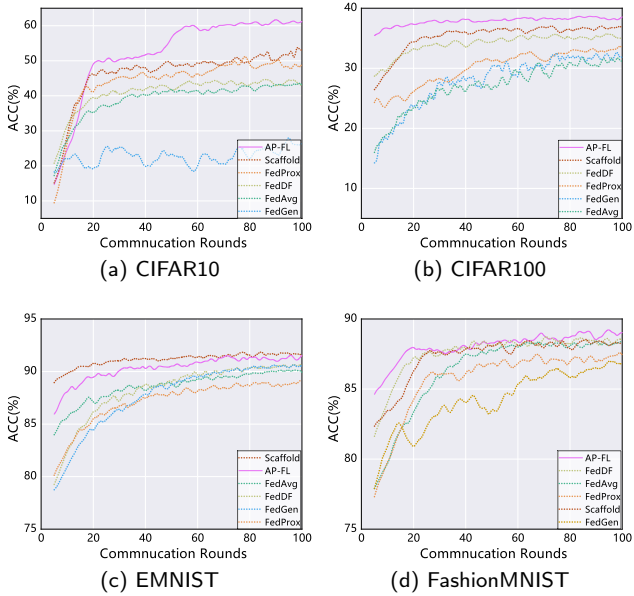

**Figure 4:** Evaluation of model performance on four datasets and five clients, the $\alpha$ set to 0.01, 0.05, 0.1 and −, respectively, for CIFAR10, CIFAR100, EMNIST, and FashionMNIST.

**Table 4**
Analysis of synthetic features on different types of semantic embedding in the dropout settings, where $\mathcal{A}_n$ corresponds to the accuracy of the friend model tested on non-dropout clients, and $\mathcal{A}_d$ corresponds to the accuracy of the friend model tested on dropout clients.

| Dataset Domain | CIFAR10 | | CIFAR100 | | EMNIST | | FashionMNIST | |
|---|---|---|---|---|---|---|---|---|
| | $\mathcal{A}_n$ | $\mathcal{A}_d$ | $\mathcal{A}_n$ | $\mathcal{A}_d$ | $\mathcal{A}_n$ | $\mathcal{A}_d$ | $\mathcal{A}_n$ | $\mathcal{A}_d$ |
| W2V | 50.74 | 41.32 | 18.92 | 15.49 | 59.86 | 44.25 | 62.14 | 50.43 |
| BERT | 55.92 | 51.63 | 21.76 | 22.05 | 65.14 | 51.27 | 73.31 | 52.84 |
| CLIP | 58.21 | 49.79 | 25.62 | 26.43 | 70.72 | 54.60 | 74.16 | 58.63 |

Regarding the number of synthetic samples, we varied the number of synthetic samples from 50 to 1000 in the experiments. As shown in Figure 5, the accuracy of the friend model on both datasets remains stable once the number of samples exceeds 600. We attribute this phenomenon to the fact that a lack of false data results in poor performance of the friend model due to the insufficient number of samples, whereas an excessive amount of false data can lead to a limited diversity of false data, which can be a bottleneck for the performance of the friend model.

### 4.3. Ablation Study

**Effect on Different Semantic Information.** In our ablation study, we investigated the impact of using different semantic embeddings in the dropout settings. Specifically, we evaluated our model with three types of semantics, namely word2vec (W2V), BERT, and CLIP. As shown in Table 4, the results with all three types of semantics are comparable, indicating the robustness of our model to different semantic embeddings. However, we observed that our model achieved the best performance with CLIP representation, suggesting the effectiveness of using CLIP as the semantic embedding.

**Effect on the Hyper-Parameters.** We performed two ablation studies on the CIFAR10 and EMNIST datasets to investigate the impact of two hyper-parameters, namely noise dimension and the number of synthetic samples, on the performance of the friend model in the Full Participation Setting. The results are presented in Figure 5. Four different noise dimensions, i.e., 20, 100, 400, and 512, were chosen to illustrate the relationship with the performance of the friend model. We observed that the performance decreases with increasing noise dimension on both datasets, indicating that high-dimensional noise may lead to significant interference.

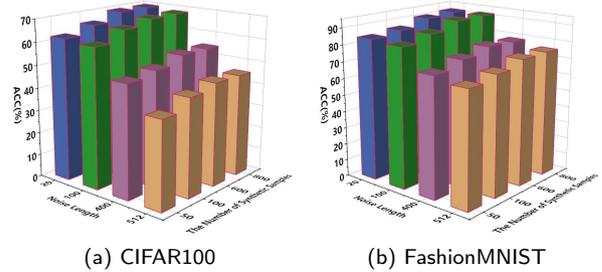

**Figure 5:** The impact of noise dimension and the number of synthetic samples on the performance of the friend model with $\alpha = 0.1$.

### 5. Conclusion

In this paper, we introduce the Asynchronous Personalized FL framework (AP-FL), which addresses the non-IID and dropout issues in FL by training a semantic generator to capture the global data distribution from non-dropout clients. This generator is then used to generate synthetic samples for each non-dropout client, aiding in the establishment of a personalized model to mitigate the client drift issue. Additionally, AP-FL leverages semantic information and the Zero-Shot learning paradigm, allowing the generator to generate previously unseen samples for dropout clients with monopoly classes and enhance data diversity for training personalized models in dropout clients. Our experiments demonstrate that AP-FL outperforms state-of-the-art methods for addressing non-IID and dropout issues in FL.